\pdfoutput=1

\documentclass[11pt]{article}

\usepackage{etoolbox}
\def\PaperType{camera}        
\ifdefstring{\PaperType}{camera}{
\usepackage{acl}
}{}
\ifdefstring{\PaperType}{review}{
\usepackage[review]{acl}
}{}

\usepackage{times}
\usepackage{latexsym}

\usepackage[T1]{fontenc}

\usepackage[utf8]{inputenc}

\usepackage{microtype}

\usepackage{bm}
\usepackage{multirow}
\usepackage{amsmath}
\usepackage{mathtools}
\usepackage[inline]{enumitem}
\usepackage{graphicx}
\usepackage{booktabs}
\usepackage{pifont}
\usepackage{mdframed}
\usepackage{dashrule}
\usepackage{caption}
\usepackage{subcaption}
\usepackage{spverbatim}
\usepackage{algorithm,algpseudocode}

\algnewcommand{\LineComment}[1]{\State \(\triangleright\) #1}

\newcommand{\matr}[1]{\mathbf{#1}}
\newcommand{\vect}[1]{\bm{#1}}
\DeclarePairedDelimiterX{\infdivx}[2]{(}{)}{%
  #1\;\delimsize\|\;#2%
}

\DeclarePairedDelimiter{\card}{\lvert}{\rvert}
\renewcommand{\Pr}{\matr{P}}
\DeclareMathOperator{\LM}{LM}

\newcommand{\JSD}{\mathrm{JSD}\infdivx}

\title{Zero-shot Generative Linguistic Steganography}

\author{
    Ke Lin\textsuperscript{1} \quad
    Yiyang Luo\textsuperscript{2} \quad
    Zijian Zhang\textsuperscript{1} \quad
    Ping Luo\textsuperscript{1} \\
    \textsuperscript{1}Tsinghua University \\
    \textsuperscript{2}Nanyang Technological University \\ 
    \texttt{\{leonard.keilin, zijianzhang510\}@gmail.com} \\
    \texttt{lawrenceluoyy@outlook.com} \\
    \texttt{luop@tsinghua.edu.cn}
}

\begin{document}
\maketitle

\begin{abstract}
Generative linguistic steganography attempts to hide secret messages into covertext. 
Previous studies have generally focused on the statistical differences between the covertext and stegotext, however, ill-formed stegotext can readily be identified by humans. 
In this paper, we propose a novel zero-shot approach based on in-context learning for linguistic steganography to achieve better perceptual and statistical imperceptibility. 
We also design several new metrics and reproducible language evaluations to measure the imperceptibility of the stegotext. 
Our experimental results indicate that our method produces $1.926\times$ more innocent and intelligible stegotext than any other method\ifdefstring{\PaperType}{review}{\footnote{The project is available in \url{https://anonymous.4open.science/r/zero-shot-GLS-DAE3}.}.}{\footnote{The project is available in \url{https://github.com/leonardodalinky/zero-shot-GLS}.}.}
\end{abstract}

\section{Introduction}

Data transmission is generally secured using encryption algorithms to create an unrecognizable ciphertext for secure data transmission \cite{bellare1997concrete}. Nonetheless, messages transmitted as ciphertexts may easily arouse suspicion from attackers, and eavesdropping may place a threat on the messages being sent. Once the communication channel is discovered, the attackers could alter the messages in any way, causing significant damage.

\begin{figure}
\centering
\includegraphics[width=\linewidth]{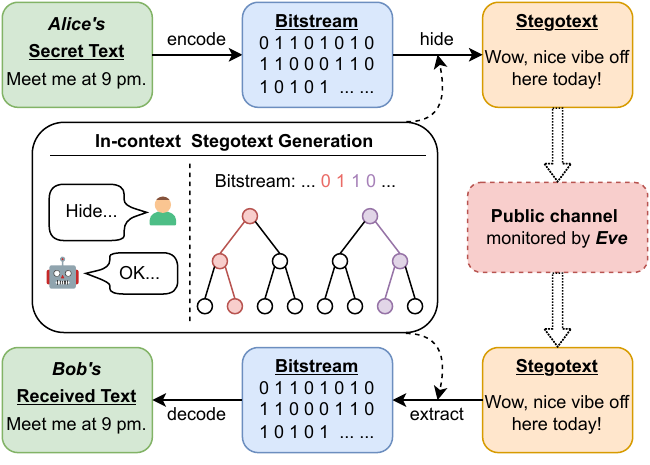}
\caption{Generative linguistic steganography pipeline.}
\label{fig:intro}
\end{figure}

Steganography is the key to ensuring communication privacy by concealing sensitive messages within monitored channels. General steganography involves embedding a secret message into a public multimedia carrier, such as text \cite{krishnan2017textoverview}, image \cite{subramanian2021imagereview,tao2019robustimagestega,guo2015jpeg}, audio \cite{lin2018rnnsm}, and video \cite{Liu2019videostegareview} to create a stego-carrier that is then transmitted via the public channel. 
As shown in Fig.~\ref{fig:intro}, the problem of secure and covert communication can be viewed as a message transmission between two parties, Alice and Bob, along with a malicious party, Eve. Eve will monitor the entire transmission and prevent unintelligible messages from being transmitted as Alice attempts to send secret information to Bob. The solution to this problem is to use a steganographic function to embed the secret message into a stego-carrier controlled by the shared secret key. The stego-carrier is then sent to Bob without alarming Eve, and Bob can extract the secret message from the stego-carrier by using the corresponding extraction function. A key objective in this scenario is to ensure that covert communications are imperceptible.


In public channels, text is commonly used, however, text contains relatively little redundant information \cite{zipf1999psycho} when compared to other media, so embedding large amounts of information can be challenging. Moreover, steganalysis tools can detect stegotext containing secret messages based on machine learning or statistical characteristics \cite{Taleby2019moderntexthiding}. To address this problem, modern steganography utilizes machine learning techniques to enhance its imperceptibility. 

Over the last few decades, researchers have devised many techniques to realize generation-based steganography \cite{xiang2022GLSreview}, including rule-based \cite{chapman1997hiding}, statistical \cite{moraldo2014markov,shniperov2016markov}, and combinations of these methods \cite{luo2016markov}. Recent developments \cite{otter2021surveyNLP} in machine learning and natural language processing have led to improved performance in linguistic steganography. It involves the use of neural networks to learn the statistical distribution of a large number of sentences, and then generate steganographic sentences from the learned language model \citep{dai2019near,yang2019tsrnn}.


Prior works have typically formulated statistical imperceptibility as the distance between the covertext distribution and stegotext distribution, however, there is no clear and simple method of measuring perceptual imperceptibility. 
Furthermore, many studies \cite{ziegler2019neural,yang2021vaestega} have demonstrated that metrics for assessing statistical imperceptibility fail to measure perceptual imperceptibility and produce inaccurate or even incorrect results. The discrepancy between the statistical and perceptual imperceptibility prevents the practical application of linguistic steganography.

\paragraph{Our Contributions.}

In this work, we propose a novel zero-shot linguistic steganography method with both improved perceptual and statistical imperceptibility. Our new method is built based on the in-context learning of large language models (LLM), which utilize some samples of the covertext to generate more intelligible stegotext using the question-answer paradigm (QA). Initially, the secret text is encoded in binary bitstreams, which are then used to generate the stegotext. At each timestep $t$, the LM computes the distribution of the $t$-th token and embeds the bitstream into the generated text by Huffman encoding. 
To demonstrate the efficacy of our proposed method compared with different supervised and training-free approaches, we develop several new metrics and conduct reproducible experiments of language evaluation.

In summary, our contributions are as follows:
\begin{itemize}
    \item 
    We present a zero-shot approach for linguistic steganography based on in-context learning using samples of the covertext.
    \item 
    We improve both the binary coding process and the embedding process by introducing several novel techniques. 
    \item 
    We design several metrics and language evaluations to evaluate both the perceptual and statistical imperceptibility, whereas our method produces more innocent and intelligible stegotext compared to all the previous methods.
\end{itemize}

\section{Background}
\subsection{Generative Linguistic Steganography}
General text generation task aims at creating token sequence $\vect{x}=[x_1,\dots,x_n]$ from the joint probability of language model $\Pr_\text{LM}(\vect{x})$. The generation process can often be viewed as an auto-regression:
\begin{equation}
\begin{aligned}
\Pr_\text{LM}(\vect{x}) &= \prod_{t=1}^n \Pr_\text{LM}(x_t\mid x_1,\dots,x_{t-1}) \\
&= \prod_{t=1}^n \Pr_\text{LM}(x_t\mid x_{<t})
\end{aligned}
\end{equation}

The goal of linguistic steganography is to send a secret message $\vect{m}\in \{0,1\}^l\subset \mathcal{M}$ to the receiver through a monitored public channel. 
In generative linguistic steganography, the sender and receiver share an embedding algorithm $f_\text{emb}: \mathcal{P}_\text{LM}\times \mathcal{M}\mapsto\mathcal{P}_\text{stega}$, which transforms $\Pr_\text{LM}$ into the steganographic distribution $\Pr_\text{stega}$ controlled by message $\vect{m}$. 
It is then used to achieve covert communication by generating \emph{stegotext} $\vect{y}\sim \Pr_\text{stega}$. 
Similarly, there exists an extraction process $f_{ext}: \mathcal{P}_\text{stega}\mapsto\mathcal{M}$ that extracts the message $\vect{m}$ from the stegotext $\vect{y}$.

\subsection{Statistical Imperceptibility}
\label{sec:statistical}
To measure the statistical imperceptibility under surveillance by an eavesdropper, \citet{cachin1998information} proposed to use the Kullback-Leibler divergence (KLD) between the distributions of natural text $\Pr_\text{true}$ and stegotext $\Pr_\text{stega}$. Since we approximate the distribution of true text $\Pr_\text{true}$ using a language model $\Pr_\text{LM}$, there is a gap between the distribution predicted by the model and the true distribution of text. Hence, we need to approximate this gap by finding an upper bound.
We can formulate the upper bound of imperceptibility by using Jensen–Shannon divergence (JSD) instead:
\begin{multline}
\JSD{\Pr_\text{true}}{\Pr_\text{stega}}^{\frac{1}{2}} \le \JSD{\Pr_\text{true}}{\Pr_\text{LM}}^{\frac{1}{2}} \\ 
+ \JSD{\Pr_\text{LM}}{\Pr_\text{stega}}^{\frac{1}{2}}
\end{multline}
Assuming that the language model is successful in modeling the covertext distribution, we can simplify the statistical imperceptibility as $\JSD{\Pr_\text{LM}}{\Pr_\text{stega}}$. 
This strong assumption can be considered true due to language models' performance.
In other words, it is feasible to measure imperceptibility using only a language model $\Pr_\text{LM}$ and a steganographic model $\Pr_\text{stega}$.

\section{Methodology}

\begin{figure*}
\centering
\includegraphics[width=\linewidth]{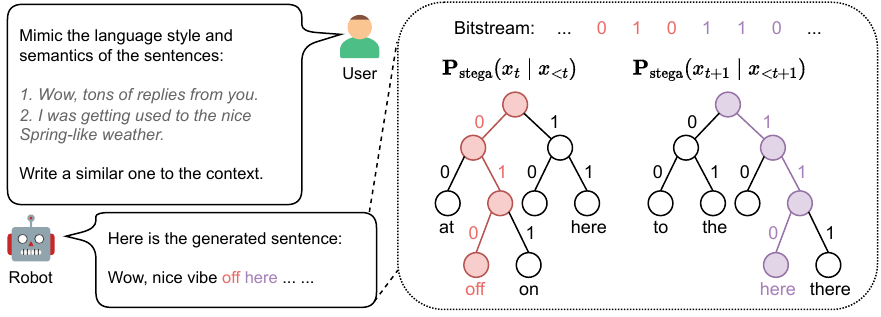}
\caption{
A running example of the embedding module and in-context stegotext generation. The selected context of samples from covertext in {\color{gray} gray} is used to instruct the stegotext generation.
 Huffman trees are constructed based on the conditional distribution, with the word matching the prefix of a bistream being selected as the next token. For example, the \textcolor[HTML]{EA6B66}{red token} has 5 candidate words after applying probability pruning. Huffman tree is constructed for the 5 candidates and the word "\textcolor[HTML]{EA6B66}{off}" matches the prefix of the bitstream and is chosen as the next word.
}
\label{fig:method}
\end{figure*}

In this section, we present the overall framework of our approach, consisting of three modules: 
\begin{enumerate*}[label=(\alph*)]
    \item \emph{Codec} module to (de)compress secret text into (from) bitstreams,
    \item \emph{Embedding} module to select candidate words from the distribution of stegotext,
    \item \emph{In-Context Stegotext Generation} module to approximate the joint distribution of covertext.
\end{enumerate*}
The pipeline of our framework is illustrated in Fig.~\ref{fig:intro}. Figure~\ref{fig:method} also explains the details of the embedding module and the in-context stegotext generation.

\subsection{Codec}
The sender in linguistic steganography wishes to send secret text to the receiver, which must first be converted into a binary bitstream. We refer to this process as encoding and decoding (codec).

\paragraph{Variable-length Coding.}
There are two categories of encoding algorithms: fixed-length coding (FLC) and variable-length coding (VLC). 
We employ Huffman coding of the VLC family for token-level encoding, as it is a relatively simple algorithm and has a sub-optimal compression ratio. Tokens are represented in binary form using the Huffman code table conditioned on $\matr{P}_\text{LM}(x_t\mid x_{<t})$.

\begin{figure}
\centering
\includegraphics[width=\linewidth]{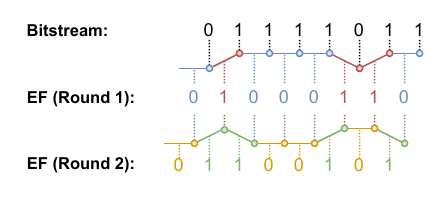}
\caption{Example of EF coding. The \textcolor[HTML]{B85450}{red} and \textcolor[HTML]{82B366}{green} lines represent that the coding in bitstream changed (from 0 to 1, or from 1 to 0) and therefore will be assigned as 1 for the next iteration. The \textcolor[HTML]{6C8EBF}{blue} and \textcolor[HTML]{D79B00}{yellow} line represents that the coding in bitstream did not change (from 0 to 0, or from 1 to 1) and therefore will be assigned as 0 for the next iteration. Bitstream is processed through several rounds until the fewest 1s exist.}
\label{fig:EF}
\end{figure}

\paragraph{Edge Flipping Coding.}
We observe that more frequent occurrences of sequential zeros in the bitstreams can improve the quality of the generated text in the embedding module due to the local ordering of the Huffman tree (i.e., the left branch always has a higher probability than the right branch). Hence, we introduce Edge Flipping coding (EF), inspired by the differential encoding of communication~\cite{weber1978differential}, on top of the Huffman process to achieve better text generation performance.

In EF coding, the bitstream is transformed to record the position at which the bits change based on the edge-triggered flip-flops in electronics. 
We apply multi-round EF coding (e.g. 16 rounds) and the bitstream with the least number of 1s is retained. A simple example of EF is provided in Fig.~\ref{fig:EF}. The analysis of EF coding is presented in Section~\ref{sec:ablation}.

\subsection{Embedding}
The goal of the embedding module is to hide and extract bitstreams within regular sentences. 

\paragraph{Hide \& Extract.}
During each phase of the generative operation, the LM yields a set of \emph{candidate words}. These words are determined based on the subsequent token's probability and can potentially serve as alternatives for the next possible word to be generated.
Then the LM will determine the length of the binary bitstream that can be embedded based on the number of candidate words.

Algorithm~\ref{alg:information_hiding} shows the brief algorithm of the hiding process. The binary representation of selected words in the conditional word distribution is obtained using the Huffman encoding algorithm. It should be noted that this Huffman process is independent of the aforementioned Huffman coding in the Codec section. Following \citet{dai2019near}, we use the threshold $\tau$ to reduce perplexity by pruning words with lower probabilities.

\begin{algorithm}
\caption{Information Hiding Algorithm}
\label{alg:information_hiding}
\begin{algorithmic}[1]
\Require Bitstream $\vect{m}$, threshold $\tau$.
\Ensure stegotext $\vect{y}=[y_1,\dots,y_n]$.

\State Timestep $t\gets 1$, output sentence $\vect{y}\gets \emptyset$
\While{not the end of $\vect{m}$}
    \LineComment{\texttt{Compute conditional probs}}
    \State $\vect{p} \gets \Pr_\text{stega} (x_t\mid x_{<t})$
    \LineComment{\texttt{Prune candidate words}}
    \State $\vect{c}\gets [c_i\mid \vect{p}(c_i) \ge \tau]$  
    \LineComment{\texttt{Huffman encoding}}
    \State $H\gets \mathrm{Huffman}(\vect{c}, \vect{p})$
    \LineComment{\texttt{Select candidate}}
    \State $y_t\gets$ Word $c\in H$ whose binary representation matches the prefix of $\vect{m}$
    \State $\vect{y}\gets \vect{y}\cup \{y_t\}, t\gets t+1$
\EndWhile
\end{algorithmic}
\end{algorithm}

As the reverse process of hiding, the information extraction process extracts steganographic information from stegotext. With a few minor differences, the extraction of secret bitstream is almost identical to the hiding process. 
The extraction algorithm is left out in Appendix~\ref{sup_sec:extraction} for simplicity.

\paragraph{Annealing Selection.}
Candidate word selection with specific thresholds $\tau$ may result in trivial candidates in some cases (e.g., $\card{\vect{c}}=1$). We note that trivial candidates often lead to a chain of trivial cases in subsequent stegotext generation, reducing the embedding rate and bit per word (BPW). 
Inspired by the stimulated annealing algorithm~\cite{vanLaarhoven1987}, we propose the annealing search to prevent trivial candidates:
\begin{equation}
\Pr'_\text{stega} (x_t\mid x_{<t}) = \mathrm{Softmax}(\frac{\LM(x_t\mid x_{<t})}{T_t})
\end{equation}
Here, the temperature $T_t$ at timestep $t$ varies according to the previous candidates $\vect{c}_{t-1}$, the initial temperature $T_0$, and the annealing factor $\alpha$:
\begin{equation}
T_t\gets
\begin{cases}
\alpha\cdot T_{t-1} & \card{\vect{c}_{t-1}}=1 \\
T_0 & \card{\vect{c}_{t-1}}>1
\end{cases}
\end{equation}

\paragraph{Repeat Penalty.}
Another corner case is that several generated words may repeat themselves after a few turns, but escape the previous annealing cases. The recently generated words are penalized for short-term repetition by applying a moving penalty:
\begin{equation}
\Pr''_\text{stega} (x_t\mid x_{<t}) = \mathrm{Softmax}(\LM(\cdot) - \vect{\delta}_t)
\end{equation}
Penalties are based on the decay factor $\beta$, initial penalty $\delta_0$, and the previously selected word. For any word $c\in \vect{c}_{t-1}$, $\vect{\delta}_t(c)$ has the following value:
\begin{equation}
\vect{\delta}_t(c)\gets
\begin{cases}
\max\{0, \vect{\delta}_{t-1}(c) - \beta\} & c\text{ not selected}\\
\delta_0 & c\text{ selected}\\
\end{cases}
\end{equation}

\subsection{In-Context Stegotext Generation}
The core of the embedding module is to calculate the conditional probability $\Pr_\text{stega} (x_t\mid x_{<t})$. We use the question-answer paradigm with pre-defined prompting context $C$ using the LLaMA2-Chat-7B \cite{touvron2023llama} as the stegotext generator. 
In other words, the distribution of stegotext can be represented as $\Pr_\text{stega} (x_t\mid x_{<t}, C)$, which is conditioned on both the previously generated tokens and the context from the covertext.

\paragraph{Context Selection.}
To approximate the distribution of covertext, we propose to incorporate several random samples from the covertext dataset into the prompting context. Since the covertext dataset is mostly available to both the sender and receiver, only a little information (e.g., random state for selection) is required to be pre-shared across different parties to reconstruct the prompting context. Therefore, an eavesdropper who suspects stegotext will be unable to decrypt the message.

\paragraph{In-Context Generation.}
As LLM is capable of comprehending the semantics of context due to its generalization ability, we utilize in-context QA tasks as a way to instruct the model to mimic the style and semantics of the covertext. 
To be more specific, we design the following structured QA format: <\texttt{ctx}$_1$,\dots,\texttt{ctx}$_k$,\texttt{[MISS]}>, where the QA model is instructed to complete the missing sentence \texttt{[MISS]} which is similar to the context. The number of contextual sentences $k$ is set as a hyper-parameter during inference. The details of the in-context QA task can be found in Appendix~\ref{sup_sec:in_context}.

\section{Experiments}

\begin{table*}
\centering
\scalebox{0.80}{
\begin{tabular}{cccccccccccc}
\toprule
\multirow{2}{*}{\textbf{Methods}} & \multirow{2}{*}{\textbf{Training-free}} & \multicolumn{5}{c}{\textbf{IMDB}} & \multicolumn{5}{c}{\textbf{Twitter}} \\
\cmidrule(lr){3-7} \cmidrule(lr){8-12} &  & BPW & PPL & JSD\textsubscript{full} & JSD\textsubscript{half} & JSD\textsubscript{zero} & BPW & PPL & JSD\textsubscript{full} & JSD\textsubscript{half} & JSD\textsubscript{zero} \\
\midrule
\multirow{3}{*}{\begin{tabular}[c]{@{}c@{}}RNN-Stega\\ (LSTM)\end{tabular}} & \multicolumn{1}{l}{\multirow{3}{*}{}} & 1.978 & 10.23 & 30.33 & 33.12 & 38.27 & 2.556 & 13.04 & 39.92 & 38.97 & 48.10 \\
 & \multicolumn{1}{l}{} & 2.682 & 12.80 & 26.76 & 29.36 & 34.87 & 3.359 & 15.38 & 36.20 & 35.53 & 44.76 \\
 & \multicolumn{1}{l}{} & 3.351 & 17.02 & 22.66 & 25.72 & 30.28 & 4.139 & 19.78 & 32.17 & 31.75 & 39.19 \\
 \midrule
\multirow{3}{*}{\begin{tabular}[c]{@{}c@{}}VAE-Stega\\ (BERT-LSTM)\end{tabular}} & \multicolumn{1}{l}{\multirow{3}{*}{}} & 1.972 & 9.68 & 34.50 & 36.47 & 38.53 & 2.247 & 10.06 & 46.07 & 45.82 & 46.61 \\
 & \multicolumn{1}{l}{} & 2.601 & 12.38 & 31.31 & 33.02 & 34.56 & 2.861 & 12.39 & 43.89 & 44.02 & 43.64 \\
 & \multicolumn{1}{l}{} & 3.199 & 16.31 & 30.03 & 31.49 & 32.82 & 3.438 & 16.13 & 42.12 & 42.54 & 40.87 \\
 \midrule
ADG & \multicolumn{1}{l}{} & 4.931 & 56.22 & 18.24 & 21.19 & 22.86 & 5.702 & 63.86 & 25.92 & 25.35 & 27.68 \\
 \midrule
\multirow{3}{*}{NLS} & \multirow{3}{*}{\ding{52}} & 1.889 & 10.40 & 23.63 & 22.83 & 17.91 & 2.059 & 10.95 & 37.71 & 36.17 & 29.34 \\
 &  & 2.531 & 12.90 & 22.08 & 21.28 & 16.73 & 2.806 & 14.01 & 36.61 & 35.17 & 29.45 \\
 &  & 3.140 & 16.70 & 20.37 & 19.63 & 14.44 & 3.513 & 18.68 & 34.42 & 32.90 & 30.18 \\
 \midrule
\multirow{3}{*}{SAAC} & \multirow{3}{*}{\ding{52}} & 4.471 & 28.74 & 18.28 & 16.40 & 13.17 & 5.078 & 36.74 & 33.75 & 32.11 & 23.08 \\
 &  & 4.749 & 37.89 & 18.04 & 16.06 & 11.49 & 5.299 & 43.35 & 33.42 & 31.82 & 22.33 \\
 &  & 5.111 & 44.02 & 17.87 & 15.98 & 11.44 & 5.716 & 54.35 & 33.20 & 31.68 & 22.04 \\
 \midrule
\multirow{3}{*}{\textbf{ours}} & \multirow{3}{*}{\ding{52}} & 1.906 & 8.81 & 17.90 & 16.86 & 13.40 & 2.550 & 9.48 & 30.90 & 29.34 & 24.90 \\
 &  & 2.420 & 13.70 & 18.37 & 17.37 & 13.67 & 3.265 & 14.44 & 30.99 & 29.45 & 25.32 \\
 &  & 3.376 & 45.04 & 18.61 & 17.87 & 13.91 & 4.029 & 47.37 & 31.74 & 30.18 & 25.40 \\
\bottomrule
\end{tabular}
}
\caption{
Experimental results under different BPWs and datasets. JSDs are multiplied by $10^2$. With the \emph{same} BPW, lower JSD $\downarrow$ indicates better performance. There is no data available for smaller BPW in ADG or SAAC due to their design of high embedding rates. 
Our proposed method shows a reduced Psic Effect in comparison to prior methods as our JSDs rise with increasing BPW, whereas other methods see a decrease in JSDs.
}
\label{tab:results}
\end{table*}

\subsection{Experiment Setup}
\paragraph{Datasets.}
Our experiments evaluate the prior works and our method using two widely used large-scale datasets, the IMDB dataset \cite{maas2011imdb} and the Twitter dataset \cite{go2009twitter}. The texts in each dataset are broken down into sentences by the \texttt{nltk} toolkit. Statistics are shown in Table~\ref{tab:statistics}.

\begin{table}
\scalebox{0.92}{
\begin{tabular}{cccc}
\toprule
\textbf{Dataset} & \textbf{\# sentences} & \textbf{Avg. Words} & \textbf{Avg. Bits} \\
\midrule
IMDB & 494,716 & 26.70 & 146.43 \\
Twitter & 2,063,307 & 12.31 & 114.85 \\
\bottomrule
\end{tabular}
}
\caption{Statistics of the preprocessed datasets.}
\label{tab:statistics}
\end{table}

\paragraph{Baselines.}
Since no training is needed for our zero-shot approach, we compare the performance of our method with several \emph{training-free} methods:
\begin{enumerate*}[label=(\alph*)]
    \item \textbf{NLS}~\cite{ziegler2019neural}: This method embeds a secret bitstream using GPT-2~\cite{radford2019gpt2} for sentence complement.
    \item \textbf{SAAC}~\cite{shen2020near}: This method improves the embedding process using Self-Adjusting Arithmetic Coding.
\end{enumerate*}

Furthermore, we also compare our benchmarks against some \emph{fully-supervised} steganographic techniques:
\begin{enumerate*}[label=(\alph*)]
    \item \textbf{RNN-Stega}~\cite{yang2019rnnstega}: This method utilizes Recurrent Neural Network models trained on covertext to generate stegotext.
    \item \textbf{VAE-Stega}~\cite{yang2021vaestega}: This method adopts a Variational Auto-encoder to generate stegotext.
    \item \textbf{ADG}~\cite{zhang2021provably}: This method proposes the Adaptive Dynamic Grouping techniques for better imperceptibility.
\end{enumerate*}

\paragraph{Implementation Details.}
All of the models and benchmarks are implemented with \texttt{PyTorch}. Pre-trained models (e.g., GPT-2 and LLaMA2-Chat) are obtained using \texttt{Huggingface} library.
To ensure a fair comparison, we rebuild all baseline methods using the same datasets throughout the entire experiment. Bits per word (BPW) are used to measure the capability of hiding information within stegotext. Only steganographic methods with 
similar BPW should be compared. Additionally, we adopt a unified BPW by the \texttt{nltk} toolkit to eliminate vocabulary inconsistencies. In each setting, 8K stegotext is generated for evaluation. 
We adopt the LLaMA2-Chat-7B as the QA model and set the context size of our method to $k=2$, the threshold $\tau=0.005$, the factors $\alpha=1.25,\beta=0.5$, and the initial values $T_0=1.0,\delta_0=4.0$. More details of baselines can be found in Appendix~\ref{sup_sec:reimplementation}.

\subsection{Evaluation Metrics}
\paragraph{Perplexity.}
As a measure of the generation quality, the perplexity (PPL) is commonly used:
\begin{equation}
\mathrm{PPL}=2^{-\frac{1}{n}\sum_{i=1}^n \log_2{\Pr(x_i\mid x_{<i})}}
\end{equation}

\paragraph{JSD.}
Following Section~\ref{sec:statistical}, we measure the statistical imperceptibility of the covertext and stegotext using JSD. Specificly, we finetune two GPT-2 models on covertext $\mathcal{C}_\text{cover}$ and stegotext $\mathcal{C}_\text{stega}$, respectively, representing $\Pr_\text{LM}$ and $\Pr_\text{stega}$. 
An evaluation of the JSD is then conducted using randomly sampled text $\mathcal{C}_\text{sample}$.
Three variants of JSD with respect to $\Pr_\text{LM}$ and $\mathcal{C}_\text{sample}$ are designed:
\begin{itemize}
    \item \emph{Full} JSD: Stego-models are trained on $\mathcal{C}_\text{cover}$ if needed\footnote{Only supervised methods need training on covertext.}. $\mathcal{C}_\text{sample}$ are sampled from $\mathcal{C}_\text{cover}$. Full JSD measures the statistical imperceptibility between covertext and stegotext.
    \item \emph{Half} JSD: Covertext dataset is split into two equal parts, $\mathcal{C}_\text{cover}^1$ and $\mathcal{C}_\text{cover}^2$. While the stego-models are trained on $\mathcal{C}_\text{cover}^1$, the evaluation is performed on $\mathcal{C}_\text{cover}^2$. It estimates imperceptibility under different but similar distributions.
    \item \emph{Zero} JSD: Zero JSD is similar to Full JSD except that it uses a vanilla GPT-2 model as the $\Pr_\text{LM}$, without fine-tuning on the covertext. This evaluates the imperceptibility between the stegotext and the normal text.
\end{itemize}

\paragraph{However, PPL and JSDs are not good metrics for assessing different stegosystems.} While PPL and JSD comparisons are suitable for supervised text generation tasks like machine translation, where a direct comparison with a golden text is possible, they face challenges in general text generation due to the absence of such direct relationships between generated output and training data. In steganography, where language distributions may vary significantly across methods, comparing PPLs and JSDs becomes impractical. Additionally, variated settings of language models, such as datasets, hyperparameters, and model architecture, will further complicate such comparisons, even if these models are within the same domain or task.

Though PPL and JSD \emph{cannot} be used to compare generation ability among different methods, it still provides an indication of performance between different settings of the same method.

\subsection{Experimental Results}

Table~\ref{tab:results} summarizes the main results on the IMDB and Twitter datasets.
Some methods are designed in such a way that it is inapplicable to adjust the BPW during sentence generation, especially for a small BPW, resulting in a lack of experiments with certain settings. This also poses a challenge for comparing metrics between similar BPWs.

\paragraph{Psic Effect} An interesting finding in our results is the Perceptual-Statistical Imperceptibility Conflict Effect (\textbf{Psic Effect}) mentioned by \citet{yang2021vaestega}, a phenomenon in which an increase in BPW and PPL is associated with a decrease in JSD. As shown in Table~\ref{tab:results}, Figure~\ref{fig:case_study}, and Table~\ref{tab:steganalysis}, while techniques like NLS exhibit superior PPL and JSD scores, they produce lower-quality text and are more susceptible to detection by attackers.
The existence of this unusual phenomenon implies that it will be more difficult for machines to detect stegotext when the text becomes more \emph{chaotic} and \emph{unintelligible}. 
This indicates the inconsistency of imperceptibility as perceived by humans and machines, as examples shown in Appendix~\ref{sup_sec:examples}. We believe that previous methods may have fallen into some local optimum of covertext distribution, but do not achieve a balance between the quality and distribution of the stegotext.

In conclusion, the state-of-the-art metrics do not match our method in every case, but our stegotext is much more natural and \textbf{does not suffer from the Psic Effect}, as is evidenced in the case study (Fig.~\ref{fig:case_study}) and Section~\ref{sec:evaluation}.

Furthermore, experiments with three variants of JSD have shown that our zero-shot method can be used to fool experts with specialized backgrounds and ordinary people without prior knowledge. 
The generalization ability of our methods ensures better performance than supervised methods in both few-shot and large-data scenarios.

\begin{figure*}
\centering
\begin{mdframed}
\small
\textbf{Secret Bitstream (Base64):} cli2EARMSajA

\hdashrule[0.5ex]{\linewidth}{1pt}{1pt}

\hspace*{\fill}\textit{Fully-supervised}\hspace*{\fill}

\textbf{RNN-Stega}: Richards are also in the film and the film shows us the most disgusting and violent aspects of the movie.The plot revolves around a group of people who are attacked and murdered.

\textbf{VAE-Stega}: listen. the music and the music, the music, music of all the actors, the music, and the music! this movie is not worth your eyes.

\textbf{ADG}: Richian is condemned to discover the facts that blackmails people through them.

\hdashrule[0.5ex]{\linewidth}{1pt}{1pt}

\hspace*{\fill}\textit{Training-free}\hspace*{\fill}

\textbf{NLS}: There was so much I wanted to see in this film. I was hoping it was a little more like what I was hoping for from the original film and not what I do not expect.

\textbf{SAAC}: I was looking forward to going as Keanu. I really wanted to do a little red carpet before I die.

\hdashrule[0.5ex]{\linewidth}{1pt}{1pt}

\textbf{Ours}: I was genuinely underwhelmed by the film, and I'm afraid the only thing that stood out to me was the final scene. I had been eagerly anticipating it, but unfortunately the rest of the movie fell flat for me.
\end{mdframed}
\caption{
Examples of stegotext generated by various methods about movie reviews with a similar $\text{BPW}\approx2.5$, except ADG and SAAC. Compared with other baseline methods, our method generates more reasonable sentences.
}
\label{fig:case_study}
\end{figure*}

\subsection{Steganalysis}

\begin{table}
\scalebox{0.85}{
\begin{tabular}{ccccc}
\toprule
\multirow{2}{*}{\textbf{Methods}} & \multicolumn{2}{c}{\textbf{Syntactic}}  & \textbf{Semantic} \\
\cmidrule(lr){2-3} \cmidrule(lr){4-4} & TS-BiRNN & R-BiLSTM-C & BERT-C     \\
\midrule
\multicolumn{4}{c}{\textit{Fully-supervised}} \\
RNN-Stega                       & 94.02   & 93.88      & 96.50    \\
VAE-Stega                       & 94.75   & 95.65     & 96.17    \\
\midrule
\multicolumn{4}{c}{\textit{Training-free}} \\
NLS                       & 84.60    & 86.21      & 92.25   \\
\textbf{ours}                      & \textbf{80.29}   & \textbf{84.34}    & \textbf{89.61}    \\
\bottomrule
\end{tabular}}
\caption{
Steganalysis with $\text{BPW}\approx 2.5$. Each value is the classification accuracy by steganalyzer. Lower accuracy $\downarrow$ indicates better statistical imperceptibility.
}
\label{tab:steganalysis}
\end{table}

To assess the statistical imperceptibility of stego-methods, a classifier is usually deployed as a steganalyzer to distinguish between covertext and stegotext.
Table~\ref{tab:steganalysis} presents the results of our steganalysis. Most prior works have primarily focused on syntactic similarity, which we believe is insufficient as text evaluation should encompass both syntactic analysis and semantic analysis.
Thus we utilize three distinct methods, namely TS-BiRNN~\cite{yang2019tsrnn}, R-BiLSTM-C~\cite{niu2019rbilstmc} and BERT-C, to assess the performance of stego-methods in terms of syntactic similarity and semantic similarity. 
Among these methods, the BERT-C method, inspired by \citet{wen2022few}, utilizes a BERT encoder \cite{jacob2019bert} to encode the input sequence and uses a linear classifier to determine whether the given sequence is a stegotext. 

We exclude ADG and SAAC from our analysis since their BPW settings are difficult to adjust, leading to an unfair comparison. Moreover, due to their high BPW settings (BPW > 5), ADG and SAAC cause significant Psic Effect, which makes it difficult to classify their stegotext accurately through syntactic analysis. Even with extremely low text quality, some of these methods can achieve almost 50\% detection accuracy, which indicates a perfect but unreasonable statistical imperceptibility. 

In conclusion, the results show that our method outperforms most of the previous methods. 

\subsection{Language Evaluation}
\label{sec:evaluation}
While statistical imperceptibility can measure the similarity between text distributions, human perception still plays a significant role in assessing the \textbf{perceptual imperceptibility} of the stegotext.

To simulate the human evaluation and ensure reproducibility, we conduct language evaluation on the generated stegotext of each method using large language models. First, we attempt to collect the stegotext generated with similar $\text{BPW}\approx2.5$ using the IMDB dataset\footnote{For ADG and SAAC, the BPW is set to 4.931 and 4.471.}. 
We then randomly select two sentences from each set of stegotext to compare, denoted by \texttt{<sent$_A$,sent$_B$>}. 
Next, GPT-3.5 is used to determine which sentence in the pair is considered high-quality. 
To be more specific, we compare our method with each baseline and collect ratios of how much of our stegotext is deemed to be better than the others. We sample 10K sentence pairs between our method and each baseline. 
Appendix~\ref{sup_sec:language_judge} illustrates the details of the implementation of language evaluation.

\begin{figure*}
\centering
\begin{subfigure}[t]{0.49\linewidth}
    \centering
    \includegraphics[width=\linewidth]{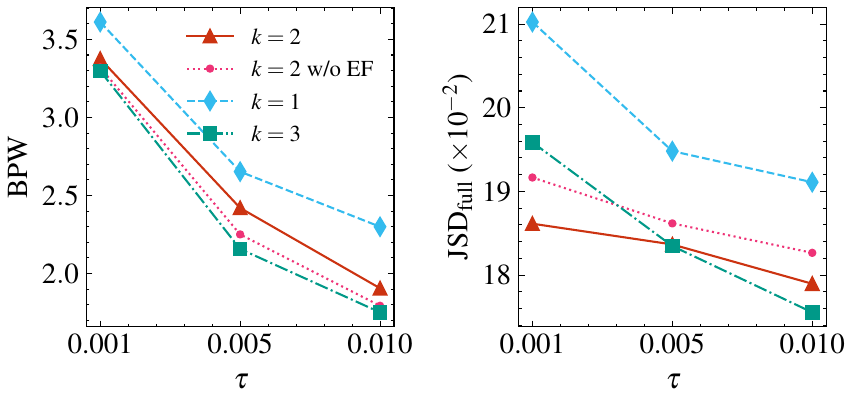}
    \caption{}
\end{subfigure}
\begin{subfigure}[t]{0.49\linewidth}
    \centering
    \includegraphics[width=\linewidth]{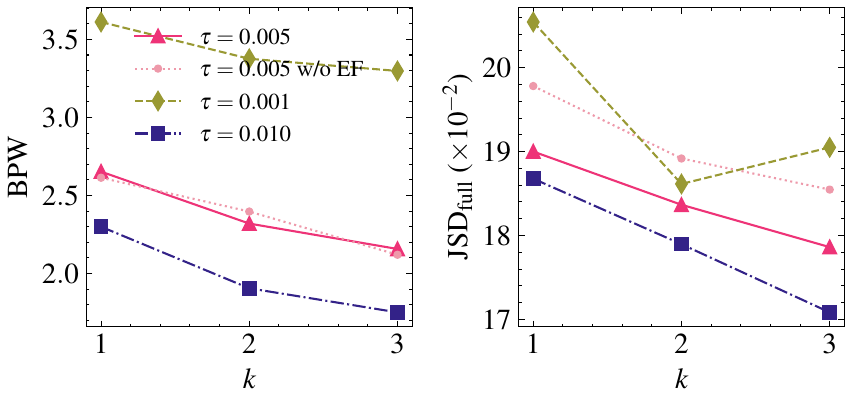}
    \caption{}
\end{subfigure}
\caption{
Curves of BPW and JSD\textsubscript{full} with respect to: (a) different threshold $\tau$ (b) different context size $k$
}
\label{fig:ablation_tau_c}
\end{figure*}

\begin{table}
\centering
\scalebox{0.8}{
\begin{tabular}{ccccccc}
\toprule
\multirow{2}{*}{\textbf{Eval.}} & \multirow{2}{*}{\textbf{GT}} & \multicolumn{3}{c}{\textbf{Fully-supervised}} & \multicolumn{2}{c}{\textbf{Training-free}} \\
\cmidrule(lr){3-5} \cmidrule(lr){6-7} &  & RNN & VAE & ADG & NLS & SAAC \\
\midrule
Sound. & 0.788 & 3.812 & 8.363 & 8.042 & 1.373 & 1.654 \\
Relev. & 1.196 & 2.397 & 3.608 & 5.345 & 2.479 & 3.850 \\
Engag. & 1.157 & 5.386 & 9.267 & 7.224 & 1.924 & 2.380 \\
\midrule
\textbf{Avg.} & \textbf{1.047} & \textbf{3.865} & \textbf{7.080} & \textbf{6.870} & \textbf{1.926} & \textbf{2.628} \\
\bottomrule
\end{tabular}
}
\caption{
Language evaluation results. Each value represents the ratio of how much our stegotext is considered better than the particular method. For example, a ratio of 3.812 means that the number of our stegotext considered better is 3.812$\times$ more than the other method.
}
\label{tab:evaluation}
\end{table}

Table~\ref{tab:evaluation} demonstrates three kinds of quality measurement conducted in our experiments:
\begin{enumerate}[label=(\alph*),itemsep=-0.5ex]
    \item \textbf{Soundness} measures the extent to which one sentence is more \emph{logical} and \emph{meaningful}.
    \item \textbf{Relevance} measures the extent to which one sentence is more \emph{relevant} to the specific topic.
    \item \textbf{Engagingness} measures the extent to which one sentence is more \emph{appealing} and \emph{engaging}.
\end{enumerate}
Our three measurements are all close to 1.0 when compared to the ground truth covertext, which indicates a high level of similarity. Most supervised methods aim to approximate covertext distribution, which results in a large gap between readability and imperceptibility. Training-free methods, however, are more effective at balancing both perceptual and statistical imperceptibility. Furthermore, our method outperforms all other baselines with an average improvement of at least $1.926\times$.

\subsection{Ablation Study}
\label{sec:ablation}

\begin{figure}
\centering
\includegraphics[width=\linewidth]{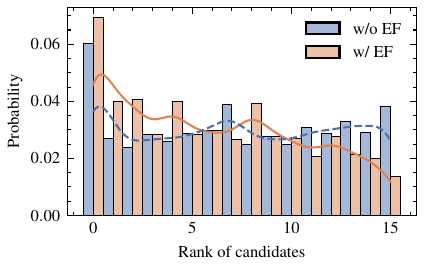}
\caption{Distribution of the rank of selected candidates.}
\label{fig:candidate_pool}
\end{figure}

We analyze EF coding in terms of the probability of selecting a candidate from a pool of candidates. We replace every occurrence of the candidate with its rank of likelihood in the descending-sorted candidate pool in Fig.~\ref{fig:candidate_pool}. It appears that words with lower ranks are more likely to be selected, which is generally of high word probability. The shift of word distribution results in slight improvements in BPW and JSDs, as shown in Fig.~\ref{fig:ablation_tau_c}.

Figure~\ref{fig:ablation_tau_c} also shows the performance of our method under different hyper-parameter settings. We can observe from the results that an increase in threshold $\tau$ and context size $k$ would result in a decrease in BPW and JSD. 
According to the results, a higher sentence quality would lead to a lower embedding rate but better statistical imperceptibility. 
This indicates a tradeoff between embedding rate and statistical imperceptibility exists.

\section{Related Work}
\label{sec:related_work}

Linguistic steganography can be divided into three types~\cite{fridrich2009steganography}, selection-based~\cite{sajedi2008coverselection,chen2019selection}, edit-based~\cite{chang2014synonym} and generation-based steganography~\cite{kang2020generative,xiang2020charlevel}. 
However, the low embedding rate of these methods hinders their practical application, giving rise to generation-based steganography.

To encode secret messages into textual data, early generative steganography utilized the Markov model to obtain conditional probability distributions \cite{moraldo2014markov} for text generation. 
Subsequently, the development of NLP has significantly improved distribution modeling. 
It was first proposed by \citet{fang2017lstm} to use a supervised recurrent neural network (RNN) to learn the statistical language model of natural texts with a fixed-length coding dictionary. Subsequently, \citet{yang2019rnnstega} proposed a generation-based steganography based on RNN and Huffman coding. Yang \textit{et al.} also proposed that different neural models make a difference, such as the generative adversarial network \cite{yang2020gantstega} and variational auto-encoder \cite{yang2021vaestega}.
\citet{zhang2021provably} and \citet{ding2023discop} further presented provably secure algorithms for better statistical imperceptibility.

While supervised methods attempt to fit models to the source distribution, the restriction of the source distribution leads to a high risk of generating similar but unintelligible text due to the complicated process of embedding secret messages.
To address this limitation, models pre-trained on large-scale datasets are used.
\citet{ziegler2019neural} proposed that arithmetic coding be used with a pre-trained GPT-2 model. 
\citet{dai2019near} further proposed an improved Huffman encoding algorithm, which is intended to adjust embedding rates dynamically.
\citet{shen2020near} improved the previous works with self-adjusting arithmetic coding for near-imperceptible text.
Nevertheless, all of these training-free approaches employ a similar paradigm for generating the context's \emph{next sentence}, making them impractical in actual usage.


\section{Conclusion}
In this work, we present a zero-shot approach based on in-context learning for linguistic steganography. For the LLM to generate a similar sentence, samples of covertext serve as the context for the LLM in the QA. We also integrate several techniques to improve the text generation process for better readability and imperceptibility of the stegotext. Our experimental results exhibit the high imperceptibility of our approach compared to other methods. In addition, the stegotext is more intelligible and resembles the covertext in many aspects. We hope this work will be a step towards the more practical applications of linguistic steganography.


\section{Limitations}
Our approach has indeed demonstrated advantages in producing intelligible and innocent stegotext, yet we rely heavily on the generalization capabilities of LLMs. The computation overhead of LLMs is approximately $3\times$ to $5\times$ in comparison with prior works. There may be an additional vulnerability if real-time communication is required.

Further, we have not tested the proposed method in a relatively large embedding rate setting (e.g., BPW > 5). While we believe that stegotext generated under settings like these would be highly unintelligible and impractical in real life, the limits of our method have yet to be determined.

Data integrity is another essential requirement of our steganographic method. The incompleteness of the stegotext may result in invertible damage to the covert bitstream as in most previous works. Fortunately, textual data are less likely to suffer from lossy compression issues than other types of media (e.g., image, audio, and video).

\section{Ethical Considerations}
The use of pre-trained LLMs may exhibit potential issues, such as insults, political biases, or gender discrimination. It should also be noted that the data underlying our method may introduce additional ethical concerns. While we did not observe such problems during our experiments, users should be aware of these issues in practice.

\ifdefstring{\PaperType}{review}{
}{
\section*{Acknowledgement}
This research is supported by the National Key R\&D Program of China under grant (No.2022YFB2703001). Thanks to all the reviewers for their insightful suggestions.
}

\bibliography{IEEEabrv,main}

\appendix

\newpage
%
%
\section{Detailed Implementation}

\subsection{Edge Flipping Coding}
The pseudo-code for Edge Flipping Coding is presented in Algorithm \ref{alg:EF_encode} for encoding, and Algorithm~\ref{alg:EF_decode} for decoding.

\begin{algorithm}
\caption{Edge Flipping Encoding}
\label{alg:EF_encode}
\begin{algorithmic}[1]
\Require Bitstream $\vect{m}$.
\Ensure EF bitstream $\vect{m}'$.

\State Output bitstream $\vect{m}'\gets \emptyset$
\State Bit state $s\gets 0$

\For{$b\in\vect{m}$}
    \If{$s = b$}
        \State $\vect{m}'\gets \vect{m}'\cup \{0\}$
    \Else
        \State $\vect{m}'\gets \vect{m}'\cup \{1\}$
    \EndIf
    \LineComment{Update bit state}
    \State $s\gets b$
\EndFor
\end{algorithmic}
\end{algorithm}

\begin{algorithm}
\caption{Edge Flipping Decoding}
\label{alg:EF_decode}
\begin{algorithmic}[1]
\Require EF Bitstream $\vect{m}'$.
\Ensure Bitstream $\vect{m}$.

\State Output bitstream $\vect{m}\gets \emptyset$
\State Bit state $s\gets 0$

\For{$b'\in\vect{m}'$}
    \If{$b' = 1$}
        \State $s\gets \mathord{\sim} s$
    \EndIf
    \LineComment{Append bit state}
    \State $\vect{m}\gets \vect{m}\cup \{s\}$
\EndFor
\end{algorithmic}
\end{algorithm}

Since EF coding is a bijection, it can be applied multi-round to the original bitstream. The final bitstream is the one with the fewest 1s.

\subsection{Bitstream Extraction}
\label{sup_sec:extraction}

We provide the bitstream extraction algorithm of the embedding module in Algorithm~\ref{alg:information_extract}.

\begin{algorithm}
\caption{Information Extraction Algorithm}
\label{alg:information_extract}
\begin{algorithmic}[1]
\Require stegotext $\vect{y}=[y_1,\dots,y_n]$, threshold $\tau$.
\Ensure Bitstream $\vect{m}$.

\State Timestep $t\gets 1$, output bitstream $\vect{m}\gets \emptyset$
\While{not the end of $\vect{m}$}
    \LineComment{\texttt{Compute conditional probs}}
    \State $\vect{p} \gets \Pr_\text{stega} (x_t\mid x_{<t})$
    \LineComment{\texttt{Prune candidate words}}
    \State $\vect{c}\gets [c_i\mid \vect{p}(c_i) \ge \tau]$  
    \LineComment{\texttt{Huffman encoding}}
    \State $H\gets \mathrm{Huffman}(\vect{c}, \vect{p})$
    \LineComment{\texttt{Select candidate}}
    \State $\vect{s}\gets$ Bits $\vect{s}$ whose binary representation matches certain word $c\in H$
    \State $\vect{m}\gets \vect{m}\cup \vect{s}, t\gets t+1$
\EndWhile
\end{algorithmic}
\end{algorithm}

\subsection{In-Context QA}
\label{sup_sec:in_context}
The complete in-context QA templates of our method are presented in Fig~\ref{fig:QA_template}. We instruct the LLM to follow the XML-style response for parsing results. Covertext context is placed in the {\color{magenta} \texttt{\{CONTEXT\}}} slot. There is also a hint of the corpus placed on the {\color{magenta} \texttt{\{CORPUS\}}} slot.

\begin{figure*}
\centering
\small
\ttfamily
\begin{mdframed}
<<SYS>>
\newline
You are an expert at mimicing the language style of others (e.g., the use of words and phrases). And you are a helpful and respectful assistant.
\newline
\newline
Users will input sentences from a given corpus. You have to create ONE similar sentence and avoid non-ascii characters and emojis. This is very important to the user's career.
\newline
\newline
The input format contains a list of sentences and where the sentences come from. For example:
\newline
<CORPUS>{\color{magenta} \texttt{\{CORPUS\}}}</CORPUS>
\newline
<CONTEXT>
\newline
Example sentence 1.
\newline
\newline
Example sentence 2.
\newline
</CONTEXT>
\newline
\newline
Your output should be like:
\newline
\newline
The generated similar sentence in ONE LINE is:
\newline
\newline
<</SYS>>
\newline
\newline
[INST]<CORPUS>{\color{magenta} \texttt{\{CORPUS\}}}</CORPUS>
\newline
<CONTEXT>
\newline
{\color{magenta} \texttt{\{CONTEXT\}}}
\newline
</CONTEXT>[/INST]
\newline
\newline
The generated similar sentence in ONE LINE is:

\end{mdframed}
\caption{
QA templates for in-context stegotext generation. Slot {\color{magenta} \texttt{\{CORPUS\}}} and {\color{magenta} \texttt{\{CONTEXT\}}} would be replaced by the corpus name and the selected context from covertext in the runtime.
}
\label{fig:QA_template}
\end{figure*}

\subsection{Baselines}
\label{sup_sec:reimplementation}
We rebuild the entire benchmark pipeline for a fair comparison between different stego-methods. The following are the implementation details:
\begin{itemize}
    \item \textbf{RNN-Stega}: We refactor the \href{https://github.com/YangzlTHU/RNN-Stega}{official codebase} from \texttt{TensorFlow} to \texttt{PyTorch}. Hyper-parameters: lr=0.001, dropout=0.5, 30 epochs, 768 LSTM units, and 2 layers of LSTM.
    \item \textbf{VAE-Stega}: We rebuild the VAE pipeline using PyTorch following the \href{https://github.com/YangzlTHU/VAE-Stega}{official codebase}.  Hyper-parameter setting: lr=3e-4, 20 epochs, hidden size of 768, DistilBERT\textsubscript{BASE} as encoder, 2 layers of LSTM as decoder.
    \item \textbf{ADG}: We reproduce ADG based on the \href{https://github.com/Mhzzzzz/ADG-steganography}{official codebase}. Hyper-parameter setting: lr=0.001, 30 epochs, 768 LSTM units, and 2 layers of LSTM.
    \item \textbf{NLS}: We reproduce NLS based on the \href{https://github.com/harvardnlp/NeuralSteganography}{official codebase}. We use \texttt{gpt2-medium} model from Huggingface Library as the base model.
    \item \textbf{SAAC}: We reproduce SAAC based on the \href{https://github.com/mickeysjm/StegaText}{official codebase}. We use \texttt{gpt2-medium} model from Huggingface Library as the base model. The hyper-parameter $\delta$ is set the same as the original paper: 0.01, 0.05 and 0.10.
\end{itemize}

\subsection{Language Evaluation}
\label{sup_sec:language_judge}
In this section, we provide the actual prompts for ChatGPT to discriminate the quality of the sentence pair \texttt{<sent$_A$,sent$_B$>}. Figure~\ref{fig:lang_eval} illustrates the prompting templates. We also randomly swap the sentences of a pair before calling the ChatGPT API.

\begin{figure*}
\small
\begin{subfigure}[b]{\textwidth}
\centering
\ttfamily
\begin{mdframed}
The user will input 2 sentences. 

You have to decide which one is more logically sound and meaningful. This is important to the user's career.
\newline
\newline
The result should be in JSON. 

It should contain the key "result" with value being either 0 or 1, indicating the first or second one.
\newline
\newline
1. {\color{magenta} <sent$_A$>}
\newline
\newline
2. {\color{magenta} <sent$_B$>}
\end{mdframed}
\caption{
Template for evaluating soundness.
}
\label{fig:soundness}
\end{subfigure}

\begin{subfigure}[b]{\textwidth}
\centering
\ttfamily
\begin{mdframed}
The user will input 2 sentences. 

You have to decide which one is more likely to be part of some "{\color{cyan} \{TOPIC\}}". This is important to the user's career.
\newline
\newline
The result should be in JSON. 

It should contain the key "result" with value being either 0 or 1, indicating the first or second one.
\newline
\newline
1. {\color{magenta} <sent$_A$>}
\newline
\newline
2. {\color{magenta} <sent$_B$>}
\end{mdframed}
\caption{
Template for evaluating relevance. Here {\color{cyan} \texttt{\{TOPIC\}}} is used to restrict the evaluation to specific topics. For IMDB, we use "movie reviews" as the topic.
}
\label{fig:relevance}
\end{subfigure}

\begin{subfigure}[b]{\textwidth}
\centering
\ttfamily
\begin{mdframed}
The user will input 2 sentences. 

You have to decide which one is more engaging and appealing to readers. This is important to the user's career.
\newline
\newline
The result should be in JSON. 

It should contain the key "result" with value being either 0 or 1, indicating the first or second one.
\newline
\newline
1. {\color{magenta} <sent$_A$>}
\newline
\newline
2. {\color{magenta} <sent$_B$>}
\end{mdframed}
\caption{
Template for evaluating engagingness.
}
\label{fig:engagingness}
\end{subfigure}
\caption{Templates for language evaluations.}
\label{fig:lang_eval}
\end{figure*}

The raw data of language evaluations is presented in Table~\ref{tab:raw_evaluation}, where each grid represents the percentage of how much ours is considered better. 

\begin{table}
\centering
\scalebox{0.8}{
\begin{tabular}{ccccccc}
\toprule
\multirow{2}{*}{\textbf{Eval.}} & \multirow{2}{*}{\textbf{GT}} & \multicolumn{3}{c}{\textbf{Fully-supervised}} & \multicolumn{2}{c}{\textbf{Training-free}} \\
\cmidrule(lr){3-5} \cmidrule(lr){6-7} &  & RNN & VAE & ADG & NLS & SAAC \\
\midrule
Sound. & .4406 & .7922 & .8932 & .8894 & .5786 & .6232 \\
Relev. & .5446 & .7056 & .7830 & .8424 & .7126 & .7938 \\
Engag. & .5364 & .8434 & .9026 & .8784 & .6580 & .7041 \\
\bottomrule
\end{tabular}
}
\caption{
Raw results of language evaluations.
}
\label{tab:raw_evaluation}
\end{table}

\subsection{Ablation}
We present the ablation study of the factors $\alpha$ and $\beta$ of annealing search and repeat penalty in Table~\ref{tab:more_ablation}.

\begin{table}
\scalebox{0.8}{
\begin{tabular}{c|ccc|ccc}
\toprule
\multirow{2}{*}{\textbf{Metrics}} & \multicolumn{3}{c|}{$\alpha$} & \multicolumn{3}{c}{$\beta$} \\
\cmidrule(lr){2-4} \cmidrule(lr){5-7} & 1.00 & 1.25 & 1.50 & 0.25 & 0.50 & 1.00 \\
\midrule
\textbf{BPW} & 2.149 & 2.420 & 2.462 & 2.116 & 2.420 & 2.511 \\
\textbf{JSD\textsubscript{full}} & 18.11 & 18.37 & 18.66 & 18.28 & 18.37 & 19.02 \\
\bottomrule
\end{tabular}
}
\caption{Ablation on the annealing search factor $\alpha$} and repeat penalty factor $\beta$.
\label{tab:more_ablation}
\end{table}

%
%
\section{Efficiency}
Table~\ref{tab:efficiency} shows the computational overhead of baselines and our method. Due to the use of LLMs, our method requires a cost of $3\times$ to $5\times$ more time.

\begin{table}
\scalebox{0.75}{
\begin{tabular}{c|cccccc}
\toprule
\textbf{Method} & \textbf{RNN} & \textbf{VAE} & \textbf{ADG} & \textbf{NLS} & \textbf{SAAC} & \textbf{ours} \\
\midrule
\textbf{Avg. Time (s)} & 1.55 & 2.16 & 1.87 & 2.89 & 3.38 & 8.41 \\
\bottomrule
\end{tabular}
}
\caption{
The average time cost of generating one steganographic sentence using a single RTX3090.
}
\label{tab:efficiency}
\end{table}

%
%
\section{Examples}
\label{sup_sec:examples}
Figure~\ref{fig:more_examples} provides more examples of the generated stegotext of each method. In comparison with other approaches, ours has better sentence quality.

\begin{figure*}
\begin{subfigure}[b]{\textwidth}
\centering
\begin{mdframed}
\small
\textbf{Secret Bitstream (Base64):} xB0CEVgADkP2YoA=

\hdashrule[0.5ex]{\linewidth}{1pt}{1pt}

\hspace*{\fill}\textit{Fully-supervised}\hspace*{\fill}

\textbf{RNN-Stega}: Sin City is a great film, and the story is told from the viewer.The plot is pretty simplistic, and has no redeemable qualities to it whatsoever.

\textbf{VAE-Stega}: short! i have the feeling that i have a copy on dvd. i would have been watching it, and i'm a huge fan.

\textbf{ADG}: Sinbad?No?So  HOAME.horrible movie.Acting wise as it describes is hard.

\hdashrule[0.5ex]{\linewidth}{1pt}{1pt}

\hspace*{\fill}\textit{Training-free}\hspace*{\fill}

\textbf{NLS}: And while the latter is certainly true, I think it is also fair to say that CBS has done a pretty good thing. The network has been a champion for science and for health.

\textbf{SAAC}: The Center for Public Integrity of West Allis, Wisconsin, and Wythe Law School of Louisville, Kentucky reported this week that cigarette makers have guities.

\hdashrule[0.5ex]{\linewidth}{1pt}{1pt}

\textbf{Ours}: All countries have an ugly side, just as ours does. It's unfair to say that only the Philippines has this issue, though. We are all capable of such shortfalls. 
\end{mdframed}
\caption{}
\end{subfigure}
\begin{subfigure}[b]{\textwidth}
\centering
\begin{mdframed}
\small
\textbf{Secret Bitstream (Base64):} pTJAjKMijGEMkA==

\hdashrule[0.5ex]{\linewidth}{1pt}{1pt}

\hspace*{\fill}\textit{Fully-supervised}\hspace*{\fill}

\textbf{RNN-Stega}: Adding insult from this review, it's not as if it is shot on the big screen.The problem was that the film was not intended.

\textbf{VAE-Stega}: four years later he're going to get to the job, and then he becomes a suspect that he was not a guy and he has a very strange and very good

\textbf{ADG}: Adding Wes Anderson is a good story,or a movie that you will get used to.In other words, it's a great

\hdashrule[0.5ex]{\linewidth}{1pt}{1pt}

\hspace*{\fill}\textit{Training-free}\hspace*{\fill}

\textbf{NLS}: The only thing missing from the film's budget is a big name star, and that will have to be one of the film-makers' two choices.

\textbf{SAAC}: The cleanroom in which the court martial took place is flanked by treetops, allowing for ...

\hdashrule[0.5ex]{\linewidth}{1pt}{1pt}

\textbf{Ours}: Cheaper doesn’t always have to equal inferior, particularly in the case of indie films, where creators are often pushed to maximize the impact of limited resources.
\end{mdframed}
\caption{}
\end{subfigure}
\begin{subfigure}[b]{\textwidth}
\centering
\begin{mdframed}
\small
\textbf{Secret Bitstream (Base64):} CYmMLJkPqMO4CHaxyEBBXrkDs5Iw

\hdashrule[0.5ex]{\linewidth}{1pt}{1pt}

\hspace*{\fill}\textit{Fully-supervised}\hspace*{\fill}

\textbf{RNN-Stega}: Extreme to the point where a film that was made was a huge waste.The film's ending is an absolute disaster, but it was very disappointing to see a man that could easily be a good thing.The film had some pretty interesting elements, and the movie was very well done, but that's not enough for it.

\textbf{VAE-Stega}: if you like "the rose", "love" (the "i have you, you're better! ")....... the ending is very funny. this is no exception. it'is so bad!... the movie is bad.. bad!!.

\textbf{ADG}: Extremely directed by Gulna Wong who brings the thought of interest for a free part of her life.Avoid this film as a film for the best of heartless Arts cinema lover.

\hdashrule[0.5ex]{\linewidth}{1pt}{1pt}

\hspace*{\fill}\textit{Training-free}\hspace*{\fill}

\textbf{NLS}: It was only a matter for him to find a wife to take care in the meantime and then the marriage will be broken up and it can be said he'll lose all his love for his beloved. I mean it makes perfect sense, it seems to make more sense that he has to leave her for the woman he loves.

\textbf{SAAC}: She wants to break things up with him. I've always seen this as perfect because it's a result of her deep love for man in this movie. However, not everybody likes watching two men break up, you know?

\hdashrule[0.5ex]{\linewidth}{1pt}{1pt}

\textbf{Ours}: Oh my gawd, stop dting married men! Like seriously, why do we keep falling into this trap over and over? Anglea basset's chracter should totall be more concerned with the fact that her husband abandoned hre for another woma than some mediocore romance flic. Overall, I would give this movie a big fat C- for wasted poasbility and a little more thought.
\end{mdframed}
\caption{}
\end{subfigure}
\caption{Examples of stegotext generated by various methods on the topic of "movie reviews".}
\label{fig:more_examples}
\end{figure*}

\section{Scientific Artifacts}
The licenses for all the baselines are listed below:
RNN-Stega (MIT License), VAE-Stega (MIT License), ADG (MIT License), NLS (MIT License), SAAC (GPL-3.0 License). 
The licenses for models are listed below:
BERT (Apache 2.0 License), GPT-2 (Modified MIT License), LLaMA2-Chat-7B (LLAMA 2 Community License).



\end{document}